\RequirePackage{fix-cm}
\documentclass[smallcondensed]{svjour3}     % onecolumn (ditto)
\smartqed  % flush right qed marks, e.g. at end of proof
\usepackage{graphicx}
\usepackage{enumerate}
\usepackage{multirow}
\usepackage{amsmath}
\usepackage{algorithm}
\usepackage{algorithmic}
\usepackage{bm}
\usepackage{amssymb}
\usepackage{marvosym}
\usepackage[numbers]{natbib}
\journalname{}

\begin{document}

\title{Adversarial Attacks for Multi-view Deep Models%\thanks{Grants or other notes
}


\author{Xuli~Sun        \and
        Shiliang~Sun\textsuperscript{\Letter}%etc.
}


\institute{Xuli~Sun \at
              School of Computer Science and Technology, East China Normal University, Shanghai 200062, China \\
              % Tel.: +123-45-678910\\
              % Fax: +123-45-678910\\
              \email{51184506038@stu.ecnu.edu.cn}           %  \\
           \and
           Shiliang~Sun \at
           	  School of Computer Science and Technology, East China Normal University, Shanghai 200062, China
           	  \at Shanghai Institute of Intelligent Science and Technology, Tongji University, Shanghai 201804, China	\\
           	  Tel.: +86-21-62233507\\ 
           	  Fax: +86-21-62232584\\
              \email{slsun@cs.ecnu.edu.cn}
}

\date{Received: date / Accepted: date}

\maketitle

\begin{abstract}
Recent work has highlighted the vulnerability of many deep machine learning models to adversarial examples. It attracts increasing attention to adversarial attacks, which can be used to evaluate the security and robustness of models before they are deployed. However, to our best knowledge, there is no specific research on the adversarial attacks for multi-view deep models.  
% fix
This paper proposes two multi-view attack strategies, two-stage attack (TSA) and end-to-end attack (ETEA). With the mild assumption that the single-view model on which the target multi-view model is based is known, we first propose the TSA strategy. The main idea of TSA is to attack the multi-view model with adversarial examples generated by attacking the associated single-view model, by which state-of-the-art single-view attack methods are directly extended to the multi-view scenario. Then we further propose the ETEA strategy when the multi-view model is provided publicly. The ETEA is applied to accomplish direct attacks on the target multi-view model, where we develop three effective multi-view attack methods. Finally, based on the fact that adversarial examples generalize well among different models, this paper takes the adversarial attack on the multi-view convolutional neural network as an example to validate that the effectiveness of the proposed multi-view attacks. Extensive experimental results demonstrate that our multi-view attack strategies are capable of attacking the multi-view deep models, and we additionally find that multi-view models are more robust than single-view models.	
\end{abstract}

\section{Introduction}
\label{intro}
Deep machine learning models have achieved great success in many fields, such as natural language processing \cite{collobert2008unified,wu2019hierarchical}, speech recognition \cite{dahl2011context,sak2014long}, computer vision \cite{krizhevsky2012imagenet}, and medical diagnosis \cite{acharya2019deep,fujita2019decision}. However, it is found that these deep machine learning models are extremely susceptible to the adversarial perturbation: the decision made by the well-trained model can be changed by adding imperceptible perturbation to the input \cite{szegedy2013intriguing,carlini2017towards}. For the exiting machine learning classifiers, even if the benign examples are added with some slight perturbations, most of them will output incorrect classification results. Moreover, though the perturbations added would be visually indistinguishable to humans, these models misclassify adversarial examples with high confidence. In other words, the model believes that the predictions are natural, and thus is very confident. Utilizing the vulnerability of deep machine learning models to adversarial examples, adversaries can easily perform malicious attacks, which poses security concerns. Furthermore, the work in \cite{kurakin2016adversarial} has shown that adversarial examples are possible in the real world by simply taking pictures of the object. When these systems, such as the driverless car system, are applied in the real world, there will be huge security risks. Consequently, the robustness of deep machine learning models against adversarial attacks has become a crucial area of research and attracted wide attention for recent years \cite{biggio2013evasion,szegedy2013intriguing,goodfellow2014explaining,tang2019cnn,barni2019transferability}.

The existence of adversarial examples poses a serious challenge to the security of machine learning models. Various kinds of methods of adversarial attack were proposed in prior work \cite{kurakin2016adversarial,papernot2016limitations,moosavi2016deepfool,narodytska2016simple,ilyas2018black,dong2018boosting,wu2019machine,Shi2019Curls,Sethi2018Data}. According to the level of adversaries' knowledge about the target model, they can be divided into two categories: white-box threat model and black-box threat model. The adversarial examples crafted by these methods do cause a large class of models to assign incorrect labels to most of them. 

However, in addition to the well-studied single-view models, more and more multi-view models have been proposed, which effectively improve performance by making full use of the information from multiple views. The multi-view models include three main categories \cite{zhao2017multi}: i) Co-training style models \cite{nigam2000analyzing,muslea2006active,sun2011robust} maximize the consistency between different views by training alternately, which were inspired by co-training \cite{blum1998combining}; ii) Co-regularization style models \cite{sun2010sparse,sun2011multi,xie2014multi} maximize the likelihood on the single view and constrain the predictions of different views to be as consistent as possible, which were achieved by adding a regularization term in the objective function; iii) Margin consistency style models \cite{mao2016soft,chao2016consensus} leverage the latent consistency among multiple views and model the margin variables on each view. Recently, some multi-view deep models were developed, such as multi-view convolutional neural network (MVCNN) \cite{su2015multi}, Show-and-Tell \cite{vinyals2015show},
Show-Attend-and-Tell \cite{xu2015show}, NeuralTalk \cite{karpathy2015deep}, and MVCRF \cite{sun2019hybrid}, which have made remarkable progress in 3D shape classification, image captioning, sequence labeling and other popular tasks. 

Although multi-view models have wide applications and superior performance,
to our best knowledge, there is no specific research on adversarial attacks for multi-view models.
% fix 
As we all know, the specific architecture and gradient information for the target model is unavailable for adversaries in the black-box setting. Black-box threat models do not differentiate between single-view and multi-view models. However, there is a difference in white-box attacks on multi-view and single-view models. Specifically, we have access to the gradient of the model to determine the direction of the adversarial perturbation in the white-box setting. For single-view models, since only one view is available, the construction of the adversarial example requires only calculating the gradient of the loss function, which is used to determine whether some value in each example should be increased or decreased. Current white-box threat models work as mentioned above and are ill-suited for multi-view models. More specifically, the existing white-box threat models implicitly assume that each view has the same impact on the resulting classification, and we refer them as the single-view attack. However, based on the consistency and complementary principle for multiple views \cite{Xu2013A}, we have to treat each view differently. Namely, to craft the adversarial example sufficient to fool the multi-view model, we have to take the partial derivative of each view separately to determine the corresponding perturbation direction. Besides, to evaluate the sensitivity of the target model performance to changes made to different views, the proposed multi-view attack also needs to achieve the separate attack for one certain view.
% fix
In this paper, we attempt to propose effective multi-view attacks against multi-view deep models.
Besides, it is an open problem whether multi-view models are more robust to adversarial examples, and thereby we make comparisons with general single-view models and evaluate the relative robustness for multi-view and single-view models. As previously mentioned, black-box threat models do not distinguish between single-view and multi-view models, therefore we focus on the white-box threat model.

\begin{figure}
  \centering
  \includegraphics[height=64mm]{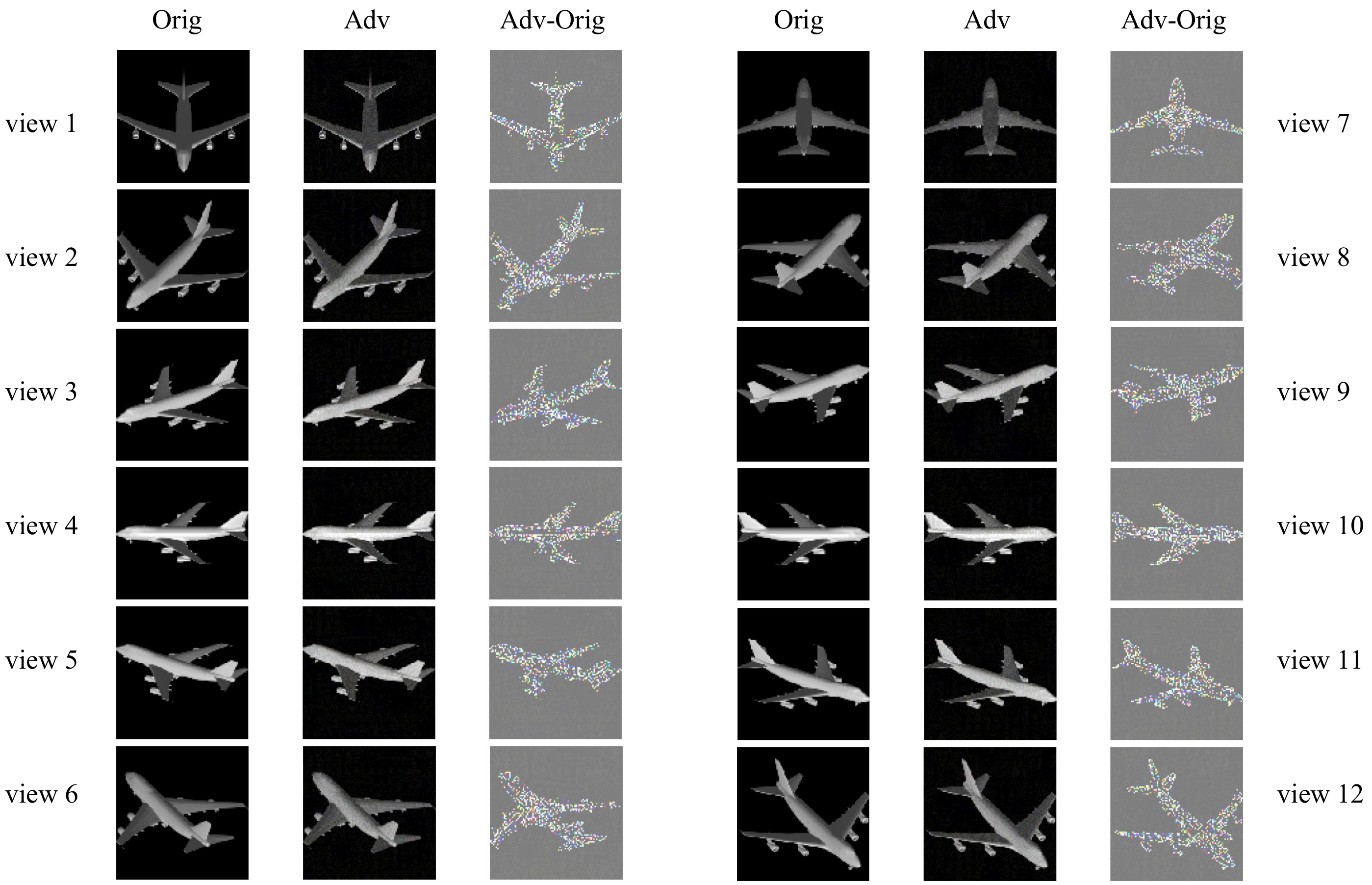}
  \caption {We show the multi-view adversarial example crafted by the proposed mBIM method. For each view, the first column is the original example, the second column is the crafted adversarial example, and the third column is the adversarial perturbation added.}
  \label{f1}
\end{figure}

Generally, adversarial attacks are specific to a certain model. However, without loss of generality, the adversarial examples crafted for one model are likely to be misclassified by others \cite{szegedy2013intriguing,liu2016delving,moosavi2017universal}. With the good transferability of adversarial examples, 
% fix
this paper takes the adversarial attack on MVCNN \cite{su2015multi} as an example to empirically study the effectiveness of the proposed multi-view attack.
We propose two attack strategies for the specific multi-view scenario: 
% fix 
i) Assuming that the underlying model on which the multi-view model is based is known, we propose the two-stage attack (TSA). Specifically, we first attack the underlying model, namely the single-view model, and then attack the target multi-view model with the crafted adversarial examples. Here, we adopt three state-of-the-art methods, fast gradient sign method (FGSM) \cite{goodfellow2014explaining}, basic iterative method (BIM) \cite{kurakin2016adversarial}, and momentum iterative method (MIM) \cite{dong2018boosting}; ii) If we have access to the target multi-view model, we then further propose the end-to-end attack (ETEA). As previously mentioned, the exiting single-view attack does not apply to attacking the target model under the multi-view scenario. We develop three effective attack methods applicable to multi-view models. The developed multi-view attack methods are referred to as mFGSM, mBIM, and mMIM, respectively, which generate multi-view adversarial examples in an end-to-end fashion. 
In order to craft imperceptible examples, we require the $L_p$ norm of the adversarial perturbation to be less than the required $\epsilon$, which is commonly exploited in adversarial attacks. As an illustration, the multi-view adversarial example crafted by the mBIM is shown in Fig. \ref{f1}.

% fix
To investigate the sensitivity of the target model performance to changes made to different views, we utilize the TSA and ETEA two strategies to first attack a single view of the model separately and then attack all views. 
Subsequently, to achieve a relatively ideal attack at the lowest budget, we adopt the ETEA strategy to attack some views of the target model greedily.
% fix
Experimental results show that multi-view models are more difficult to attack than single-view models. Specifically, the fooling rates of the multi-view model is strictly lower than the single-view models. Besides, the multi-view model cannot be successfully attacked when only one single view is perturbed; only simultaneous attacks on all or some valued views can improve the fooling rate of the target multi-view model to some extent. 
% fix 
However, for the single-view model, all attacks can cause a dramatic drop in model performance.
Moreover, the experimental results fully demonstrate that ETEA can make more effective attacks for multi-view models, whether it is the single-view attack or the all-view attack. Especially, the ultimate fooling rate for all-view attacks by exploiting the ETEA strategy is more than 60\%\footnote{The average classification accuracy of MVCNN for original testing examples is 94.5\%. After using the ETEA strategy for adversarial attacks, the accuracy rate decreases to about 30\%.}.  

The main contributions of this paper are as follows:
\begin{itemize}
\item We propose two strategies for the adversarial attacks on multi-view deep models, referred to as the two-stage attack (TSA) and end-to-end attack (ETEA).
% fix
\item We develop three multi-view attack methods, named mFGSM, mBIM, and mMIM, which have been experimentally proved to be effective against multi-view models. 
% fix  
\item We are the first to show that multi-view models are more adversarially robust than single-view models. 
\end{itemize}

The remainder of this paper is organized as follows. In Section \ref{Related Work}, we review some relevant work. Section \ref{Background} introduces some problem formulation and primary models on which our work is built. Section \ref{Methodology} elaborates on the proposed TSA and ETEA  strategies for the attacks of the multi-view models. 
% fix
Section \ref{Experiments} reports extensive experimental results that confirm the effectiveness of the proposed multi-view attacks. 
Finally, we summarize this paper and discuss possible future work in Section \ref{Conclusion}. 

\section{Related Work}
\label{Related Work}
Machine learning models are challenged by adversarial attacks. Their high vulnerability has attracted widespread attention, and the security research of models has become an active topic \cite{dalvi2004adversarial,jia2017adversarial,carlini2018audio}. Although it is often mistaken that the vulnerability to adversarial attacks is the weakness of deep models, the same problem exists on traditional machine learning models \cite{biggio2013security,biggio2014pattern,fawzi2018adversarial}. We restrict ourselves to the work of adversarial attacks on deep machine learning models in this paper. 

Regarding the adversarial vulnerability for deep machine learning models, there are the following speculative explanations: i) The model is trained by the finite data set and has incomplete generalization \cite{bengio2009learning}; ii) The model consists of highly linear components for facilitating training \cite{goodfellow2014explaining}. The existing attack methods can be divided into two categories: optimization-based methods and gradient-based methods. For the optimization-based methods, the box-constrained L-BFGS \cite{szegedy2013intriguing} is one of the most representative methods. Given an original example $x$, the search for the corresponding adversarial example $x_{adv}$ can be formalized as follows:
\begin{alignat}{2}
\mathop{\arg\min}_{x_{adv}} \quad & || x_{adv} - x ||_2 \nonumber\\
\mbox{s.t.} \quad &f(x_{adv})=\mathrm{y}^*, \,x_{adv}\in [0,1]^m\nonumber,
\end{alignat}
where $\mathrm{y^*}$ is the desired label. However, this problem is hard to solve for deep models due to their non-convexity and non-linearity \cite{larochelle2009exploring}. Therefore, the authors converted the search for minimum  perturbations into the following optimization problem: 
\begin{alignat}{2}
\mathop{\arg \min}_{x_{adv}} \quad &\lambda||x_{adv}-x||_2+J(x_{adv},\mathrm{y}^*) \nonumber \\
\mbox{s.t.} \quad &x_{adv} \in [0,1]^m\nonumber,
\end{alignat}
where $\lambda$ is the trade-off parameter and $J(\cdot)$ is a loss function.
This kind of methods directly optimize the $L_2$ distance between the original example $x$ and the adversarial example $x_{adv}$, and their main drawbacks are high computational cost and quite slow.
Besides, there are many gradient-based methods for adversarial attacks. FGSM was first proposed in \cite{goodfellow2014explaining}, which constructed $x_{adv}$ by linearizing the $L_\infty$ neighborhood of $x$. However, compared with L-BFGS, FGSM attack has a lower success rate. For the purposes of making a nice tradeoff between attack success rate and computational cost, an iterative variant of FGSM, known as BIM or I-FGSM, was introduced in \cite{kurakin2016adversarial}. Subsequently, Dong et al. \cite{dong2018boosting} further integrated momentum into the iterative process. They showed that the proposed MIM attack improved the success rate and transferability of the generated adversarial examples. 
Simply put, gradient-based methods transform the process of constructing $x_{adv}$ from $x$ into the following optimization problem:
\begin{alignat}{2}
\mathop{\arg\max}_{x_{adv}} \quad &J(x_{adv},\mathrm{y}_{true}) \nonumber \\
\mbox{s.t.} \quad &||x_{adv}-x||_\infty \leq \epsilon, \nonumber
\end{alignat}
where $\mathrm{y}_{true}$ is the true label for $x$, and $\epsilon$ is the size of adversarial perturbation. In simple terms, they use the calculated gradient to continuously update the input so that it can be eventually misclassified by the model, instead of updating the model parameters.
In short, the aforementioned methods use backpropagation to transform the construction of adversarial examples into an optimization problem based on the loss function. 

Adversarial attacks pose potential security risks on machine learning models in practical applications. More importantly, the work by Szegedy et al. \cite{szegedy2013intriguing} reported that adversarial examples could generalize across models. That is, adversarial examples designed for the model $M_1$ are often likely to fool the model $M_2$. It was also found that adversarial transferability works in real-world scenarios \cite{papernot2016transferability,papernot2017practical}. Tram{\`e}r et al. \cite{tramer2017space} showed that there was a significant overlap in the adversarial subspace between different models by empirically, and considered that it was a source of adversarial transferability. Recent work considered the phenomenon of adversarial transferability from a theoretical standpoint \cite{charles2018geometric}. 

Based on that the same adversarial examples can be misclassified with high probability by different models, we focus on the adversarial attacks on the state-of-the-art MVCNN model \cite{su2015multi}, which is a multi-view model for 3D shape classification. Unlike previous work mentioned above, 
% fix
this paper proposes attack methods applicable for multi-view models and takes the adversarial attacks against MVCNN as an example to show the effectiveness of proposed multi-view attacks. We additionally evaluate the relative robustness of multi-view models and single-view
models against adversarial examples.

\section{Problem Formulation and Primary Models}
This section first defines the problem to be solved, then describes the threat model, and finally gives a brief introduction to the multi-view target model attacked in this paper.
\label{Background}
\subsection{Problem Description}
Consider an original example $x \in \mathbf{X} $ with ground-truth label $\mathrm{\mathrm{y}_{true}} \in \mathbf{Y}$. The deep machine learning model is denoted by $\Psi$, which outputs a predictive label $\mathrm{y}_{pred}$ for $x$. Adversarial attacks aim to find an example $x_{adv} = x + \eta(x)$ to fool the target model, where $\eta(x)$ is the minor perturbation added to the $x$. Formally, the desired $x_{adv}$ needs to meet the following two requirements:
\begin{enumerate}[1)]
  \item \textbf{Imperceptible:} $x_{adv}$ looks so similar to the original $x$ that people can not tell them apart visually. Generally, the $L_p$ norm of the perturbation is required to not exceed the allowed constant $\epsilon$, $||\eta(x)||_p\leq\epsilon$, where $p \in \{0, 1, 2, \infty\}$. The $L_p$ norm is formulated as follows: 
  \begin{equation}
  ||\eta(x)||_p = \Bigg(\sum_{k=1}^{K}|\eta(x)_k|^p\Bigg)^{\frac{1}{p}}\nonumber,
  \end{equation}
  where $k$ is the number of coordinates.
  In particular, $L_\infty$ norm is considered as the optimal metric \cite{warde201611}, and thus the adversarial methods in this paper use the $L_\infty$ norm to quantify the similarity between the original example $x$ and the adversarial example $x_{adv}$.

  \item \textbf{Missclassification:} At the beginning, the target model $\Psi$ can predict correct label for $x$, i.e., $\Psi(x) = \mathrm{y_{true}}$. However, after making a small change to $x$, $\Psi$ alters the output to any class label that is different from the ground-truth label $\mathrm{\mathrm{y}_{true}}$. Specifically, $\mathrm{y_{pred}}=\Psi(x_{adv})$ and $\mathrm{y_{pred}} \ne \mathrm{y_{true}}$. In other words, our goal is to craft the adversarial example $x_{adv}$, which forces $\Psi$ to predict an incorrect label, and it is not a matter of concern which class the incorrect label falls into.
\end{enumerate}

Note that the adversarial examples are constructed in the process of testing. That is, the adversarial attacks are directly performed on the well-trained model. Any attack on the training phase is beyond our consideration.

\subsection{Threat Model}
According to the varying amount of knowledge about the target model $\Psi$ by adversaries, the scenarios for adversarial attacks can be divided into the white-box threat model and black-box threat model. Specifically, the 
\emph{white-box threat model} refers to that the adversaries have full knowledge about $\Psi$, including the model architecture, training parameters, etc. In particular, the model gradient is available under the white-box setting, and the adversary can effectively craft adversarial examples with gradient descent \cite{szegedy2013intriguing,nguyen2015deep,carlini2017towards,madry2017towards}. It should be emphasized that the computed gradient by the target model is used to update the examples rather than the model parameters. In principle, the perturbation added to the original example $x$ should be indistinguishable to humans, while $\Psi$ misclassifies it with high confidence. Generally, we can adopt early stopping to control the perturbation norm so that it satisfies the constraint of $||\eta(x)||_p\leq\epsilon$ \cite{goodfellow2014explaining,kurakin2016adversarial}, or optimize the norm by adding it to the objective function as a regularization term \cite{carlini2017towards}. When zero knowledge about $\Psi$ is known, it belongs to the \emph{black-box threat model}. Although it may not know any internal details of the model, the adversary is allowed to query $\Psi$. In other words, the adversary can feed the input to $\Psi$ and observe the output, which is the predicted class label or the confidence score that the input belongs to each category. 

% fix 
Although some effective black-box threat models have been proposed \cite{Shi2019Curls,Sethi2018Data}, we focus on the white-box threat models rather than the black-box threat models, since the latter does not distinguish between multi-view and single-view models. That is, the related work for black-box models is complementary and independent of ours. We aim to propose effective multi-view attacks for multi-view deep models.

\subsection{Target Model}
Considering that adversarial examples have good transferability among different models, this paper carries out adversarial attacks on the multi-view convolutional neural network (MVCNN) \cite{su2015multi}. The MVCNN is a state-of-the-art multi-view model for 3D shape classification, and 
% fix
we take it as an example to investigate the effectiveness of the proposed multi-view attacks. Specifically, the MVCNN model we adopted is based on the standard VGG11 network\footnote{Different convolutional neural network (CNN) architectures can be used for MVCNN, such as ResNet-M \cite{he2016deep}. However, we find experimentally that these models have comparable performance. Therefore, this paper adopts the common version, the MVCNN based on the VGG11.},
which is composed of five convolutional (Conv) layers, three fully connected (FC) layers, and a softmax classification layer. MVCNN takes the rendered images from multiple views as input. More specifically, the images on each view pass through the Conv layers of VGG11 independently, and all the parameters are shared at this stage. Then the extracted features from multiple views are aggregated by the ``view-pooling'' layer, and fed to the remaining non-linear layers for shape classification. Note that the ``view-pooling'' layer is added before FC1 (the first FC layer) and after Conv5 (the last Conv layer), where the element-wise maximum operation is performed to fuse the features across different views.

% fix
To further compare the relative robustness of the multi-view model and single-view model against adversarial examples, we make adversarial attacks on baseline single-view models with the white-box threat models used in the two-stage attack (TSA) strategy introduced in Section \ref{Two-Stage Attack}. We refer to these single-view models as SVCNN in this paper. 
Note that the SVCNN models are based on the VGG11 pre-trained on ImageNet and further fine-tune by a single view or simple concatenation of multiple views. The training process for MVCNN and the SVCNN models are introduced in Section \ref{Train}.

\section{Methodology}
\label{Methodology}
In this section, we elaborate on the two proposed attack strategies for multi-view models: two-stage attack (TSA) and end-to-end attack (ETEA). 
It has been shown that a considerable part of the adversarial examples crafted for one model can effectively fool others, sometimes with a success rate of up to 100\% \cite{papernot2016transferability,liu2016delving}. That is, the adversarial examples may generalize among different models, which is also known as adversarial transferability.
% fix
Therefore, to investigate the effectiveness of the proposed multi-view attacks, we take the above MVCNN model as an example and perform adversarial attacks on it.

\subsection{Two-Stage Attack}
\label{Two-Stage Attack}
% fix
Assume that we do not have direct access to the target multi-view model, the single-view model however on which the target model is based is available for us. 
With the mild assumption, we first propose the two-stage attack (TSA) strategy. The main idea of TSA is to first attack the single-view model on which the target multi-view model is based, and then attack the multi-view model with the generated examples.
% fix
By TSA strategy, we can compare the relative robustness of the multi-view model and single-view model and directly extend the existing attack methods onto the multi-view scenario. 
The first stage adopts three state-of-the-art attack methods to craft adversarial examples for the single-view model. They are fast gradient sign method (FGSM), basic iterative methods (BIM), and momentum iterative method (MIM), all of which belong to the gradient-based methods. 
The three methods are used for non-target attacks in this paper, as is done in most work. Recall that the non-target attack refers to making the target model predict any incorrect labels for the adversarial example. The following is a brief review of the three attack methods.

\subsubsection{Fast Gradient Sign Method (FGSM)} FGSM was proposed by Goodfellow et al. \cite{goodfellow2014explaining}. It crafts the optimal adversarial example $x_{adv}$ by maximizing the loss function.
FGSM works by using the gradient sign to estimate the perturbation that satisfies the $L_{\infty}$ norm bound $||\eta(x)||_{\infty}\leq\epsilon$. Here, $\epsilon$ is the max value of allowed adversarial perturbation. Formally, the perturbation for the original example $x$ is written as follows:
\begin{equation}
\eta(x) = \epsilon \cdot sign\Big(\nabla_x J(x, \mathrm{y}_{true})\Big),
\end{equation}
where $x$ represents the input image, $\mathrm{y}_{true}$ is the ground-truth label for $x$, and the function $J(\cdot)$ denotes the loss function. We adopt the commonly used cross-entropy loss in this paper. Note that the parameters of the model are intentionally omitted here because our adversarial attacks are performed on the well-trained model.

Based on the above setting, the corresponding adversarial example for $x$ can be obtained as follows:
\begin{eqnarray}
  x_{adv} = x + \eta(x).
\end{eqnarray}

\subsubsection{Basic Iterative Methods (BIM)}
BIM introduced in \cite{kurakin2016adversarial} intuitively extends FGSM. It increases the value of the loss function by moving multiple times along the direction of the gradient sign with smaller step size $\alpha$, instead of moving one step with step size $\epsilon$. Here, $\alpha=\frac{\epsilon}{T}$, and $T$ is the number of iteration steps. Moreover, it takes the clip operation for the updated result at each step, by which restricting the generated example in the $\epsilon$ vicinity of $x$. Formally, begin by setting $x_{adv_{(0)}} = x$, then for each iteration,
\begin{gather}
	x_{adv(t)} = Clip_{x, \epsilon}\Big\{x_{adv(t-1)} + \eta(x_{adv(t-1)})\Big\},
\end{gather}
where $x_{adv(t-1)}$ and $x_{adv(t)}$ are the adversarial examples in the $t-1$ iteration and the $t$ iteration, respectively. Furthermore, 
\begin{eqnarray}
  \eta(x_{adv(t-1)})=\alpha\cdot sign\Big(\nabla_x J(x_{adv(t-1)}, \mathrm{y}_{true})\Big),
\end{eqnarray}
is the perturbation added to the $x_{adv(t-1)}$. After $T$ iterations, we can obtain the final adversarial example $x_{adv} = x_{adv(T)}$.

\subsubsection{Momentum Iterative Method (MIM)}
MIM is a variant of BIM, inspired by the momentum method \cite{polyak1964some}, which can escape the local maximum by remembering previous gradients. MIM integrates the momentum term into the iterative procedure to generate adversarial samples. Let $T$ be the total number of iteration steps. Then the generation of the adversarial sample in the $t$ iteration by MIM is as follows:
\begin{eqnarray}
  x_{adv(t)} = x_{adv(t-1)}+ \alpha\cdot sign(g_{t}),
\end{eqnarray}
where $\alpha=\epsilon/T$ is the step size, and the accumulated gradient $g_{t}$ can be written as 
\begin{eqnarray}
  g_{t} = \mu \cdot g_{t-1} + \frac{\nabla_x J(x_{adv(t-1)},\mathrm{y}_{true})}{||\nabla_x J(x_{adv(t-1)},\mathrm{y}_{true})||_1}.
\end{eqnarray}
Here, $\mu$ represents the decay factor.

As mentioned before, 
% fix
the TSA strategy assumes the gradient of the multi-view model is not directly obtainable, but the single-view model however on which the target multi-view model is based is known. TSA first directly attacks the corresponding single-view model to generate adversarial examples for each view, and then these examples are used for the multi-view model to achieve the attack, by which to evaluate the sensitivity of the model to changes made to each view.  
Specifically, the first stage focuses on exploiting the gradient information to guide the discovery of the adversarial direction and then crafted adversarial examples, which are very similar to original examples but can fool the single-view model SVCNN. The second stage then uses the crafted adversarial examples to attack the MVCNN.

The TSA with BIM attack is summarized in Algorithm \ref{TSA}.

\begin{algorithm}[th]
   \caption{Two-Stage Attack (TSA) with BIM}
   \label{TSA}
   \begin{flushleft}
 $\#$ Stage 1: Single-view model attack.\\
 % \hspace*{0.02in}
 {\bf Input:} Well-trained single-view model $\Psi_v$ with the loss function $J(\cdot)$, original example set for the $v$-th view $\mathbf{X}^v$, true label set $\mathbf{Y}$, 
 the allowed value $\epsilon$, and the number of iteration in the first stage $T_1$.\\
{\bf Output:} Adversarial example set for the $v$-th view $\mathbf{X}_{adv}^v$. \\
 \end{flushleft}
 \begin{algorithmic}[1]
 \STATE{Initialize  $\mathbf{X}_{adv}^v = \varnothing$ and step size $\alpha=\frac{\epsilon}{T_1}$;}
 \FOR{each $x^v \in \mathbf{X}^v$}
   \STATE{Initialize $x_{adv{(0)}}^v = x^v$;}
   \FOR{$t_1=1$ to $T_1$}
   \STATE{Input $x_{adv(t_1-1)}^v$ to $\Psi_v$;}
   \STATE{Compute the perturbation for $x_{adv(t_1-1)}^v$ by applying the gradient sign as
   \begin{eqnarray}
   \eta(x_{adv(t_1-1)}^v)=\alpha\cdot sign\Big(\nabla_{x^v} J(x_{adv(t_1-1)}^v, \mathrm{y}_{true}^v)\Big)\nonumber;
   \end{eqnarray}
   }
   \STATE {Update the adversarial example as
   \begin{eqnarray}
   x_{adv(t_1)}^v = Clip_{x, \epsilon}\Big\{x_{adv(t_1-1)}^v + \eta(x_{adv(t_1-1)}^v)\Big\}\nonumber;
   \end{eqnarray}
   }\\
   \ENDFOR
   \STATE{$x_{adv}^v=x_{adv(T_1)}^v$ and add $x_{adv}^v$  to $\mathbf{X}_{adv}^v$;}\\
\ENDFOR
\STATE{Return $\mathbf{X}_{adv}^v$ with each $x_{adv}^v$ satisfies $||\eta(x^v)||_{\infty}\leq\epsilon$.}

\vspace{0.5cm}
\end{algorithmic}
\begin{flushleft}

 $\#$ Stage 2: Multi-view model attack.\\
  {\bf Input: } Adversarial example set $\mathbf{X}_{adv}^v$, original example set $\mathbf{X}^*=\mathbf{X} \verb|\| \{x^v\}$ that excluding the original examples on the $v$-th view, true label set $\mathbf{Y}$, and the number of iteration in the second stage $T_2$.\\
  {\bf Output:} Average classification accuracy ${Acc}$.\\
  \end{flushleft}
 \begin{algorithmic}[1]
 \STATE{Initialize $Acc=0$;}
   \FOR{$t_2=1$ to $T_2$}
   \STATE{Input $\{\mathbf{X}_{adv}^v,\mathbf{X}^*\}$ to the multi-view model $\Psi$;}\\
   \STATE{Obtain the $t_2$-th classification accuracy $Acc_{t_2}$;}\\
   \STATE{$Acc \leftarrow Acc+Acc_{t_2}$;}
   \ENDFOR
 \STATE{Return ${Acc}=\frac{Acc}{T_2}.$}
 \end{algorithmic}
\end{algorithm}

\subsection{End-to-End Attack}
% fix
Though the existing single-view attack methods have been proven to be simple and effective \cite{szegedy2013intriguing,carlini2017towards,madry2017towards}, as discussed previously, they are not suitable for multi-view models. To directly attack the target multi-view model, we further propose an end-to-end attack (ETEA) strategy, in which three effective multi-view attack methods are developed, called mFGSM, mBIM, and mMIM. The main idea of ETEA is to use the multi-view attack methods to attack the target model in an end-to-end fashion. 
For clarity, this paper focuses on the mBIM attack in our discussion about ETEA. We show the experimental results for all attack methods in Section~\ref{Experiments}.

Given an input-label pair $(x^1, x^2, \cdots, x^V, \mathrm{y}_{true})$, where $x^1$ denotes the original example on the first view and $x^2$ indicates the original example on the second view. For clarity, we denote the original example from multiple views as $x$, and the set of original examples is denoted by $\mathbf{X}$, i.e., $x=\{x^1, x^2, \cdots, x^V\} \in\mathbf{X}$. More specifically, $\mathrm{y}_{true}$ is the corresponding true label for $x$, and $V$ represents the total number of views of the example, whose value depends on the specific data set. For a given $x$ with the ground-truth label $\mathrm{y}_{true}$, the cross-entropy loss applied to integer class labels is equal to the negative logarithm probability of the true class label given $x$, and this relationship can be formalized as follows:
\begin{equation}
J(x,\mathrm{y}_{true})=-\mathrm{log}\,p_\Psi(\mathrm{y}_{true}|x)\nonumber.
\end{equation}
To make the multi-view model $\Psi$ predict an incorrect label for the adversarial example $x_{adv}$, i.e., $\Psi(x_{adv}) \ne \mathrm{y}_{true}$, 
we need to minimize the probability of correctly classifying $x_{adv}$. It can be written as follows:
\begin{equation}
\mathop{\max}_{x_{adv}} J(x_{adv},\mathrm{y}_{true}) = \mathop{\min}_{\mathrm{y}_{true}\in\mathbf{Y}} \mathrm{log}\,p_\Psi(\mathrm{y}_{true}|x_{adv}),
\end{equation}
where $p_\Psi(\mathrm{y}_{true}|x_{adv})$ represents the confidence level that the model $\Psi$ divides $x_{adv}$ into class $\mathrm{y}_{true}$, and $\mathbf{Y}$ is the set of class labels. Similarly, the set of adversarial examples is denoted as $\mathbf{X}_{adv}$, i.e., $x_{adv}\in\mathbf{X}_{adv}$.

Based on the above setting, we can find an imperceptible $x_{adv}$ by solving the following problem:
% aim to
\begin{alignat}{2}
\mathop{\arg\max}_{x_{adv}} \quad &J(x_{adv},\mathrm{y}_{true}) \nonumber \\
\mbox{s.t.} \quad &||\eta(x^v)||_\infty \leq \epsilon, \, v \in \{1,2,\cdots, V\},
\end{alignat}
where $\eta(x^v)$ is the adversarial perturbation added to the original example on the $v$-th view. 

Specifically, supposing we attack the $v$-th view of the example, mBIM requires an iterative procedure to craft an adversarial example $x_{adv}^v$. In more detail, $x^v$ takes a step size $\alpha$ in the direction of $\mathrm{sign}\{\frac{ \partial{J}}{\partial{x^v}}\}$ for each step, and then we clip and update the intermediate results. To begin, we set
\begin{gather}
	x_{adv_{(0)}}^v = x^v\label{eq9},
\end{gather}
and then clip $x_{adv(i)}^v$ on each iteration:
\begin{gather}
	x_{adv(i)}^v = Clip_{x^v, \epsilon}\Bigg\{x_{adv(i-1)}^v + \alpha\cdot sign\Big(\frac{ \partial{J_{i-1}}}{\partial{x^v}}\Big)\Bigg\}\label{eq10},
\end{gather}
where the loss for the $i-1$ iteration
\begin{gather}
J_{i-1}=J(x_{adv(i-1)}^v, \mathrm{y}_{true})\nonumber.
\end{gather}
More formally, $Clip\{\cdot\} $ is the clipping function, and its exact form is as follows: 
\begin{equation}
\begin{split}
Clip_{x^v, \epsilon}\{x_{adv(i)}^v\}&=\mathop{\min}\Big\{255, x^v+\epsilon,\\&\mathop{\max}\{0, x^v-\epsilon,x_{adv(i)}^v\}\Big\}, \nonumber
\end{split}
\end{equation}
which is used to ensure the resulting adversarial example is in the $L_\infty$ $\epsilon-$ neighborhood of $x^v$.
The ETEA with mBIM attack is summarized in Algorithm \ref{ETEA}.
\begin{algorithm}[th]
   \caption{End-to-End Attack (ETEA) with mBIM}
   \label{ETEA}
   \begin{flushleft}
 % \hspace*{0.02in}
 {\bf Input:} Well-trained multi-view model $\Psi$ with loss function $J(\cdot)$, original example set $\mathbf{X}$, true label set $\mathbf{Y}$, the allowed value $\epsilon$, and the total number of iteration $T$.\\
  {\bf Output:} Average classification accuracy ${Acc}$.\\
  \end{flushleft}
 \begin{algorithmic}[1]
 \STATE{Initialize $Acc=0$, adversarial example set $\mathbf{X}_{adv}=\varnothing$;}
   \FOR{each $x \in \mathbf{X}$}
   \STATE{Initialize $x_{adv}$ according to the Eq. (\ref{eq9});}
   \STATE{Update $x_{adv}$ according to the Eq. (\ref{eq10});}
   \IF{Attack successfully}
   \STATE{Add $x_{adv}$ to $\mathbf{X}_{adv}$;}
   \ELSE
   \STATE{Add $x$ to $\mathbf{X}_{adv}$;}
   \ENDIF
   \ENDFOR
   \STATE{Input $\mathbf{X}_{adv}$ to $\Psi$;}
   \FOR{$t=1$ to $T$}
   \STATE{Obtain the $t$-th classification accuracy $Acc_{t}$;}\\
   \STATE{$Acc \leftarrow Acc+Acc_{t}$;}
   \ENDFOR
 \STATE{return ${Acc}=\frac{Acc}{T}.$}
 \end{algorithmic}
\end{algorithm}

\section{Experiments}
\label{Experiments}
In this section, 
% fix
we perform adversarial attacks on MVCNN to evaluate the effectiveness of the proposed multi-view attacks, and compare the relative robustness of MVCNN and baseline single-view models.

\subsection{Benchmark}
Following \cite{su2015multi}, all our analysis is performed on the widely-used ModelNet40 data set,
which is a shape classification benchmark and consists of 40 categories. In all experiments, we adopt the standard data division as \cite{wu20153d}. Specifically, the training set and test set contain 9483 and 2468 shapes from 40 common categories, respectively. Assume that each of the 3D shapes in the data set is placed vertically (parallel to the z-axis). A virtual camera is placed every 30 degrees in the vertical radial position to generate the 2D images for 3D shapes from multiple views. It can also be imagined as there is a virtual camera that takes a picture every time the shape rotates 30 degrees in the vertical direction. Then the 2D images are rendered using Phong shading \cite{phong1975illumination}. That is, each 3D shape has different 2D shaded images from twelve views. After data augmentation, the data set contains 147732 (12311 $\times$ 12) rendered images, and the size of each image is 224$\times$224.
% fix
According to the good generalization of the adversarial examples, the model trained by different data sets can still be fooled by the same adversarial samples \cite{szegedy2013intriguing,liu2016delving,moosavi2017universal}. That is different data sets have no significant influence on the effect of the adversarial attack. Moreover, considering this ModelNet40 is publicly available, and therefore this paper reports the performance of MVCNN on this data set to demonstrate the effectiveness of the proposed multi-view attacks.

\subsection{Evaluation Metrics}
This paper adopts two evaluation metrics common for the adversarial attacks: the classification accuracy and fooling rate. Specifically, in the experiments for the adversarial attacks on single-view models, we show the average classification accuracy before and after perturbations, ${Acc}_{orig}$ vs ${Acc}_{adv}$, to demonstrate the general vulnerability for single-view models. Formally,
\begin{equation}
Acc = \frac{1}{T}\cdot{\sum_{t=1}^{T}{\sum_{j=1}^{N}C_t(x_j, \mathrm{y_{true}})}},
\end{equation}
where $\mathrm{y_{true}}$ represents the ground-truth label for the original image, $N$ is the number of testing examples, and $T$ is the number of iterations. The function $C(\cdot)$ is an indicator function, and the formal definition for it can be written as follows:
$$C(x, \mathrm{y_{true}})=
\begin{cases}
0,& \text{if $\mathrm{y_{pred}}\neq\mathrm{y_{true}}$}\\
1,& \text{otherwise.}
\end{cases}$$
Here, $\mathrm{y_{prec}}$ is the class label predicted by the model. In short, this function achieves value 1 if and only if $x$ can be classified correctly into the ground-truth category. 

The adversarial attacks on the target models may not be successful. That is, the model may still be able to correctly classify the adversarial examples generated by different methods. 
% fix
Therefore, we adopt the average fooling rate (FR) to show the results of adversarial attacks on the multi-view model and single-view models, which is used to make a more intuitive comparison of their relative robustness. 
Formally, 
\begin{equation}
FR = {Acc}_{orig} - {Acc}_{adv},
\end{equation}
where ${Acc}_{orig}$ is the average classification accuracy of the model on the original examples, and ${Acc}_{adv}$ is the average classification accuracy on the generated adversarial examples. We can intuitively see the degradation extent of the model performance with FR.

\subsection{Setup}
\label{Train}
\subsubsection{Model Training} 
Our all adversarial attacks are on well-trained models, and any attack on the training procedure is not considered in this paper. The model training procedure is as follows:
\begin{itemize}
\item{We train individually a single-view CNN (\textbf{SVCNN}) model with the rendered images from each view, which is based on the VGG11 pre-trained on ImageNet. For example, during SVCNN1 training, the images on the first view are input to VGG11. After 30 epochs, there are twelve well-trained SVCNN models, we refer them to SVCNN1, SVCNN2, $\cdots$, SVCNN12.}
% fix
Note that we further train the thirteenth single-view model SVCNN13, which is trained with a high-dimensional view formed by the concatenation of all views.
\item{\textbf{MVCNN} training needs to go through two stages: i) The images on a single view are successively input into VGG11, which is pre-trained on ImageNet, to fine-tune the model parameters. It can be regarded as the general image classification task; ii) The view-pooling layer is combined to jointly classify the images from multiple views. The training procedures for two stages are both set to 30 epochs, and the batch sizes are 64 and 96 for the first stage and the second stage, respectively. The training for MVCNN can be finished using the publicly available implementation\footnote{Different versions of implementation have been released by the authors' Lab. In this paper, we adopt the PyTorch implementation}.}
\end{itemize}

Note that all SVCNN models are trained separately with the corresponding single view.
% fix
In particular, SVCNN13 is trained with a single view formed by simple concatenation of all views. 
Moreover, to make a fair comparison, MVCNN and thirteen single-view models are trained using Adam optimizer \cite{kingma2014adam} with the initial learning rate $5\times10^{-5}$, and the weight decay is 0.001.

\subsubsection{Adversarial Attacks} 
% fix
To fully demonstrate the effectiveness of the proposed multi-view attacks, we run a series of experiments. Besides, to compare the relative robustness of the multi-view model and single-view models, we first attack SVCNN models by exploiting the state-of-the-art attack methods, FGSM, BIM, and MIM\footnote{The code of these methods is publicly available at https://github.com/baidu/AdvBox.}. 
Then the proposed TSA and ETEA are used to achieve the adversarial attacks on MVCNN. 
% fix
In particular, we attempt to adopt two strategies to attack MVCNN, respectively. We first attack only one of the views to evaluate the sensitivity of the model to changes made to each view, and then attack all views simultaneously.  
Furthermore, based on the better attack performance achieved by ETEA, we attack some important views in a greedy fashion, hoping to achieve an effective attack with the as little budget as possible. 
% fix
For all attack methods, as in most prior work, we adopt the common $L_\infty$ norm, and the maximal iterative step is set to 100. It should be noted that the magnitude of the adversarial perturbation must not be too small. For example, all attacks cannot succeed at $\epsilon$=0.01. Without loss of generality, the larger the perturbation added, the more likely the attacks to succeed. However, we find that $\epsilon$=0.1 is sufficient to complete the adversarial attack. Meanwhile, the perturbation of this magnitude is imperceptible for humans. Therefore, we do not explore it too much in this paper.

\subsection{Results and Discussion}
In this section, we first present the results of adversarial attacks on SVCNN models in Table \ref{table-1}. Then we report the corresponding attack results for MVCNN, which are presented in Tables \ref{table-2}, \ref{table-3}, and Fig. \ref{f2}, respectively.

% fix：标题，增加FR列
\subsubsection{SVCNN Attack Results}
\begin{table}
    \caption{The results of adversarial attacks on SVCNN models. ${Acc}_{orig}$ and ${Acc}_{adv}$ are the average classification accuracies on original examples and adversarial examples. FR is the average fooling rate. Column ${Acc}_{adv}$ and column FR from left to right are the results of adversarial attacks with FGSM, BIM, and MIM.}
    \label{table-1}
    \vskip 0.15in
    \centering
    \begin{small}
    \begin{tabular*}{10.8cm}{l|c|c|c|c|c|c|c}
      \hline
		\multirow{3}{*}{Model} & \multirow{3}{*}{${Acc}_{orig}$ (\%)} & \multicolumn{3}{c|}{${Acc}_{adv}$ (\%)} & \multicolumn{3}{c}{FR (\%)}\\[2pt]
		                       & & FGSM & BIM & MIM & FGSM & BIM & MIM  \\[2pt] \hline \hline
		\multicolumn{1}{l|}{SVCNN1} & 92.75 & 25.49 & 25.47 & 26.82 & 67.26 & 67.28 & 65.93 \\[1.5pt]
		\multicolumn{1}{l|}{SVCNN2} & 92.53 & 29.09 & 29.10 & 29.90 & 63.44 & 63.43 & 62.63 \\[1.5pt]
	    \multicolumn{1}{l|}{SVCNN3} & 92.64 & 33.11 & 32.84 & 34.73 & 59.53 & 59.80 & 57.91 \\[1.5pt]
	    \multicolumn{1}{l|}{SVCNN4} & 92.01 & 27.88 & 27.68 & 29.41 & 64.13 & 64.33 & 62.60 \\[1.5pt]
	    \multicolumn{1}{l|}{SVCNN5} & 92.04 & 32.83 & 32.91 & 33.65 & 59.21 & 59.13 & 58.39 \\[1.5pt]
	    \multicolumn{1}{l|}{SVCNN6} & 91.55 & 30.76 & 30.54 & 32.11 & 60.79 & 61.01 & 59.44 \\[1.5pt]
	    \multicolumn{1}{l|}{SVCNN7} & 91.77 & 25.75 & 25.76 & 27.27 & 66.02 & 66.01 & 64.50 \\[1.5pt]
	    \multicolumn{1}{l|}{SVCNN8} & 91.83 & 25.72 & 27.87 & 28.84 & 66.11 & 63.96 & 62.99 \\[1.5pt]
	    \multicolumn{1}{l|}{SVCNN9} & 92.33 & 30.84 & 31.06 & 32.80 & 61.49 & 61.27 & 59.53 \\[1.5pt]
	    \multicolumn{1}{l|}{SVCNN10} & 91.63 & 27.26 & 27.28 & 29.01 & 64.37 & 64.35 & 62.62 \\[1.5pt]
	    \multicolumn{1}{l|}{SVCNN11} & 92.53 & 31.96 & 32.21 & 34.23 & 60.57 & 60.32 & 58.30 \\[1.5pt]
	    \multicolumn{1}{l|}{SVCNN12} & 92.35 & 31.50 & 31.39 & 32.70 & 60.85 & 60.96 & 59.65 \\[1.5pt]
	    \multicolumn{1}{l|}{SVCNN13} & 92.42 & 23.96 & 23.84 & 25.02 & 68.46 & 68.58 & 67.40 \\[1.5pt]
	      \hline
    \end{tabular*}
    \end{small}
  \end{table}
We first evaluate the adversarial robustness of SVCNN models including SVCNN1, SVCNN2, $\cdots$ SVCNN13, which have been trained with original images from the associated view. Formally, we adopt FGSM, BIM, and MIM methods to attack the above thirteen single-view models, respectively. 
% fix (FR)
The average fooling rate and average classification accuracies of SVCNN models on the original and adversarial images are demonstrated in Table \ref{table-1}. 
We can see that all SVCNN models obtain excellent performance on the shape classification task: the original average accuracy is more than $90\%$ for each model. However, they are highly vulnerable to adversarial attacks. The average classification accuracy for each SVCNN model is less than $35\%$ on crafted adversarial images. 
% fix
In particular, we can see from Table \ref{table-1} that although SVCNN1 achieves the highest accuracy on the original examples, it also has a very high FR. That is, the performance of SVCNN1 is the worst after the attack, except for SVCNN13. This phenomenon suggests that view 1 may contain more useful information, and therefore it is more effective to attack this view. To verify this idea we attack each view separately in the adversarial attack on MVCNN. Besides, we also refer to original examples and find that view 1 is the front of the object. The further the other views deviate from the front, the more difficult it is to attack the corresponding single-view model trained with those views. That is, the single-view model has smaller FR. This suggests that the probability of success against an attack is related to the view being attacked. For SVCNN13, it achieves the highest FR compared to the other 12 single-view models. Although SVCNN13 leverages data from multiple views (simple concatenation), its architecture is still a generic single-view model. Random perturbations of certain dimensions can misclassify the model, which makes it easier to attack.
Results in Table \ref{table-1} show that FGSM and BIM attacks achieve comparable performance, and are more effective than MIM under the same parameter settings. The experimental results strongly demonstrate the vulnerability of single-view models.

\subsubsection{TSA Results}
\begin{table}
    \caption{The results of two-stage adversarial attacks on MVCNN, using the adversarial examples crafted for SVCNN by FGSM, BIM, and MIM.}
    \label{table-2}
    \vskip 0.15in
    \centering
    \begin{small}
    \begin{tabular*}{5.8cm}{l|c|c|c}
      \hline
      \multirow{3}*{Attacked-view} & \multicolumn{3}{c}{FR (\%)}\\[2pt]
       & FGSM & BIM & MIM\\[2pt]
      \hline 
      \hline
      view 1 & 0.25 & 0.27 & 0.25 \\[1.5pt]
      view 2 & 0.22 & 0.20 & 0.22 \\[1.5pt]
      view 3 & 0.16 & 0.16 & 0.17 \\[1.5pt]
      view 4 & 0.40 & 0.41 & 0.40 \\[1.5pt]
      view 5 & 0.16 & 0.23 & 0.22 \\[1.5pt]
      view 6 & 0.12 & 0.14 & 0.14 \\[1.5pt]
      view 7 & 0.29 & 0.30 & 0.29 \\[1.5pt]
      view 8 & 0.13 & 0.11 & 0.10 \\[1.5pt]
      view 9 & 0.11 & 0.18 & 0.16 \\[1.5pt]
      view 10 & 0.34 & 0.33 & 0.31 \\[1.5pt]
      view 11 & 0.19 & 0.21 & 0.18 \\[1.5pt]
      view 12 & 0.13 & 0.11 & 0.16 \\[1.5pt]
      \hline
      All views 1 & 7.77 & 7.75 & 8.35\\[1.5pt]
      All views 2 & 29.27 & 31.10 & 27.75\\[1.5pt]  %with the adv examples of MVCNN13
      \hline
    \end{tabular*}
    \end{small}
  \end{table}
In the experiments of adversarial attacks on MVCNN by TSA strategy, we attempt to only attack a single view at first. It can be seen from Table \ref{table-2} that the maximum fooling rate is 0.41\%, which indicates that MVCNN have certain robustness against the adversarial images. Then we attack all views by two kinds of adversarial images: one is the adversarial images generated by attacking SVCNN13 (corresponding to the \textsl{All views 1} in Table \ref{table-2}), and the other is the adversarial image set for SVCNN1-12, i.e., $\{\mathbf{X}_{adv}^1, \mathbf{X}_{adv}^2, \cdots, \mathbf{X}_{adv}^{12}\}$ (corresponding to the \textsl{All views 2} in Table \ref{table-2}). 

From Table \ref{table-2}, it can be seen that attacking each view alone has different effects on the performance of MVCNN. That is, the attack on a certain view may be more effective than others, which indicates that each view has different importance for the model to correctly classify. 
Moreover, it is clear that the classification performance of MVCNN can be significantly reduced when attacking all views. 
% fix
Specifically, if only one of the views is attacked, other views can still help the model make better predictions due to the consistency and complementary principle among multiple views. The attacks on all views can be very successful. This shows multi-view models themselves have a certain degree of adversarial robustness. However, as shown in Table \ref{table-1}, all attacks on the single-view model can cause a dramatic drop in model performance. This demonstrates indirectly that multi-view models are more robust than single-view models.
% fix
Further, it can be found that the lowest FR of single-view models in Table \ref{table-1} is close to 58\%, while the highest FR of the multi-view model in Table \ref{table-2} is only 31\%. Comparing the FR of MVCNN in Table \ref{table-2} with SVCNN models in Table \ref{table-1}, we can conclude that the multi-view model is indeed more robust to adversarial examples.
% no fix
Table \ref{table-2} also presents that \textsl{All views 1} attack is not as effective as \textsl{All views 2} attack: the latter's FR reaches more than 31\%, while the former's FR is less than 10\%. 
% fix
Each view contains different information. Without loss of generality, a specific attack on each view is more effective than the random perturbation of a high-dimensional view formed by the concatenation of all views. Therefore, we can find that the FR of all views 1 is less than all views 2. 
Further, as we can see that the adversarial images generated by MIM are superior to FGSM and BIM in \textsl{All views 1} attack, and BIM outperforms the other two methods in \textsl{All views 2} attack.

\subsubsection{ETEA Results}
\begin{figure}
  \centering
  \includegraphics[height=70mm]{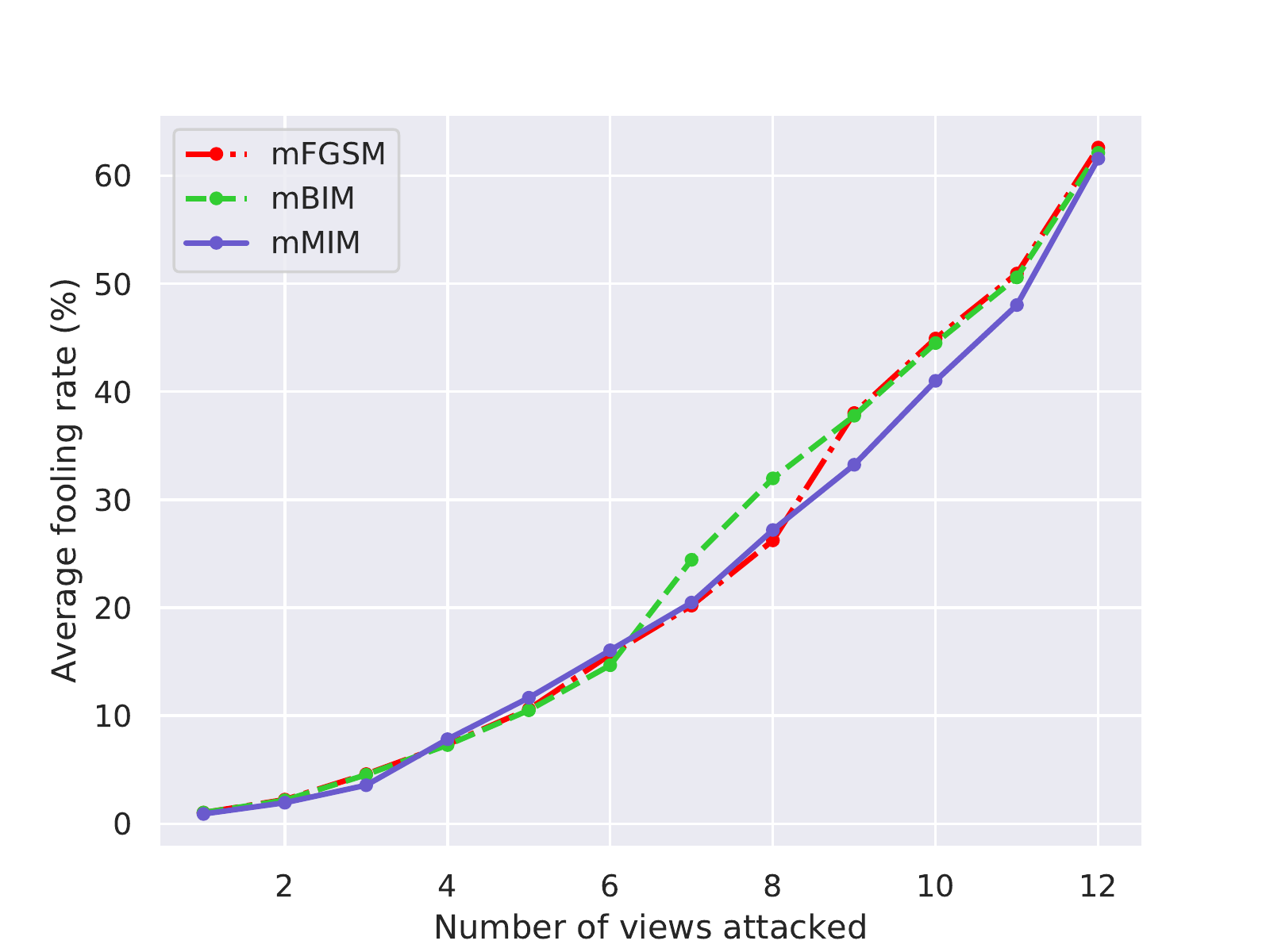}
  \caption {The results of adversarial attacks on some views of MVCNN with mFGSM, mBIM, and mMIM.}
  \label{f2}
\end{figure}

We use mFGSM, mBIM, and mMIM methods to attack MVCNN with ETEA strategy, respectively. Table \ref{table-3} reports the corresponding results, which include the cases that only a single view is attacked and all views are attacked. All experimental results indicate the effectiveness of the developed methods: the average fooling rates for the three kinds of attacks all exceed 60\%. Similarly, the attack is frustrated when we attack only a single view, and when all views are attacked can the model classification performance be significantly reduced. 
% fix
As mentioned above, the failure of single-view attacks indicates that the multi-view model itself has certain adversarial robustness, which is mainly attributed to the consistency and complementary principle among multiple views.
Further, based on different effects of single-view attacks, we attack some relatively important views greedily, hoping to achieve an effective attack with the as little budget as possible, as shown in Fig. \ref{f2}. 
We believe that finding the optimal permutation to accomplish the some-view attack is an important problem, and we will leave it for future work without going much exploration in this paper.
% fix
In other words, our multi-view attack methods are general. They can be applied whether multiple views are directly known from the feature space or not. That is, if some of the views cannot be available or lost, we still can perform the some-view attack.
It should be emphasized that there is a trade-off between the attack performance and budget. In the case of a limited budget, we can consider attacking just some of the views instead, i.e., some-view attack. For example, Fig. \ref{f2} shows that the fooling rate is 20\% or higher when seven important views are attacked. Otherwise, we can choose to attack all views to achieve the optimal attack effect.

\begin{table}
    \caption{The results of adversarial attacks on MVCNN with mFGSM, mBIM, and mMIM.}
    \label{table-3}
    \vskip 0.15in
    \centering
    \begin{small}
    \begin{tabular*}{6.4cm}{l|c|c|c}
      \hline
      \multirow{3}*{Attacked-view} & \multicolumn{3}{c}{FR (\%)}\\[2pt]
       & mFGSM & mBIM & mMIM\\[2pt]
      \hline
      \hline
      view 1 & 0.73 & 0.78 & 0.71 \\[1.5pt]
      view 2 & 0.53 & 0.56 & 0.58 \\[1.5pt]
      view 3 & 0.54 & 0.52 & 0.45 \\[1.5pt]
      view 4 & 0.80 & 0.84 & 0.79 \\[1.5pt]
      view 5 & 0.43 & 0.40 & 0.43 \\[1.5pt]
      view 6 & 0.31 & 0.40 & 0.31 \\[1.5pt]
      view 7 & 0.53 & 0.54 & 0.58 \\[1.5pt]
      view 8 & 0.55 & 0.59 & 0.55 \\[1.5pt]
      view 9 & 0.58 & 0.65 & 0.55 \\[1.5pt]
      view 10 & 1.02 & 1.03 & 0.90 \\[1.5pt]
      view 11 & 0.55 & 0.53 & 0.48 \\[1.5pt]
      view 12 & 0.24 & 0.23 &0.25 \\[1.5pt]
      \hline
      All views & 62.60 & 62.11 & 61.57\\[1.5pt]
      \hline
    \end{tabular*}
    \end{small}
  \end{table}

In summary, in the adversarial attacks against MVCNN, all single-view attacks have failed, and attacks on all views can be very successful, which indicates that 
% fix
multi-view models themselves have a certain degree of adversarial robustness. Moreover, it can be found that multi-view models are more robust by comparing the FR of MVCNN and the SVCNN models. 
Moreover, the attacks under the ETEA strategy are more effective than those under the TSA strategy, whether single-view attacks or all-view attacks. 
%fix
This well shows the effectiveness of the developed multi-view attacks.
In addition, since the attack on a certain view may be more effective, we can consider the some-view attack at a limited budget.

\section{Conclusion}
\label{Conclusion}
In this paper, we propose the TSA and ETEA strategies to attack multiple-view models. 
% fix
In particular, we develop three multi-view attack methods, called mFGSM, mBIM, and mMIM, which can be used to directly attack multi-view models in an end-to-end fashion. Experimental results demonstrate the effectiveness of the proposed multi-view attacks, and we additionally show that multi-view models are more robust for adversarial attacks, compared to the single-view models.

In this paper, we have considered the problem of adversarial attacks on multi-view models. In future work, we will focus on the defense task and propose more effective defense algorithms for multi-view models to further enhance the adversarial robustness of the models.

\begin{acknowledgements}
This work is supported by the National Natural Science Foundation of China under Project 61673179, Shanghai Knowledge Service Platform Project (No. ZF1213), and the Strategic Priority Research Program of ECNU. Prof. Shiliang Sun is the corresponding author of this paper.
\end{acknowledgements}


\bibliographystyle{spbasic}      % basic style, author-year citations
\bibliography{ref}   % name your BibTeX data base

% \begin{thebibliography}{}
% %
% % and use \bibitem to create references. Consult the Instructions
% % for authors for reference list style.
% % Format for Journal Reference
% Author, Article title, Journal, Volume, page numbers (year)
% Author, Book title, page numbers. Publisher, place (year)
% \end{thebibliography}

\end{document}